\documentclass{article}

\usepackage{microtype}
\usepackage{graphicx}
\usepackage{subcaption} 
\usepackage{booktabs} 

\usepackage{xurl}

\usepackage{hyperref}
\usepackage{float}
\usepackage{placeins}
\usepackage{dblfloatfix} 
\graphicspath{{./}} 
\newcommand{\paperfigwidth}{0.68\linewidth}
\newcommand{\figHbig}{0.30\textheight}   
\newcommand{\paperfigwidthGeom}{0.748\linewidth}
\newcommand{\figHgeom}{0.33\textheight}

\usepackage[preprint]{icml2026}

\usepackage{amsmath}
\usepackage{amssymb}
\usepackage{mathtools}
\usepackage{amsthm}

\usepackage[capitalize,noabbrev]{cleveref}
\usepackage{algorithm}
\usepackage{algorithmic}

\theoremstyle{plain}
\newtheorem{theorem}{Theorem}[section]

\theoremstyle{definition}

\theoremstyle{remark}
\newtheorem{remark}[theorem]{Remark}

\usepackage[disable,textsize=tiny]{todonotes}
\usepackage{xcolor}
\usepackage[normalem]{ulem}
\makeatletter
\long\def\new#1{{#1}}
\makeatother

\icmltitlerunning{PIEFS: Physics-Informed Eigenfunction Features}

\begin{document}

\twocolumn[
  \icmltitle{PIEFS: Physics-Informed Eigenfunction Features with Learnable Scaling}



  \icmlsetsymbol{equal}{*}

  \begin{icmlauthorlist}
    \icmlauthor{Varvara Nazarenko}{HSE}
    \icmlauthor{Timur Lidzhiev}{HSE}
    \icmlauthor{Alexander Tarakanov}{HSE,AIVK}
  \end{icmlauthorlist}

  \icmlaffiliation{HSE}{Faculty of Computer Science, HSE University, Moscow, Russia}
  \icmlaffiliation{AIVK}{AI VK, Moscow, Russia}

  \icmlcorrespondingauthor{Varvara Nazarenko}{varunaza@gmail.com}
  \icmlcorrespondingauthor{Timur Lidzhiev}{trlidzhiev@gmail.com}
  \icmlcorrespondingauthor{Alexander Tarakanov}{atarakanov@hse.ru}

  \icmlkeywords{PIEFS, Physics-Informed Machine Learning, Artificial Intelligence for Science, Dirichlet Energy, Representation Learning, Spectral Methods, ICML}

  \vskip 0.3in
]



\printAffiliationsAndNotice{\textit{Accepted to the AI4Physics Workshop at ICML 2026.}}  

\begin{abstract}
Spectral methods are widely used to construct representations from the geometry of data, but they often rely on a fixed kernel, graph Laplacian, or manually selected feature scaling. We propose Physics-Informed Eigenfunction Features with Learnable Scaling (PIEFS), a \new{supervised neural representation-learning framework with a spectral inductive bias}, based on a modified Dirichlet energy. In PIEFS, scalar coordinate maps are trained under empirical Gram orthogonality, a supervised linear readout, and a Dirichlet penalty in which the input gradient is transformed by a learnable metric $A(x)=\Lambda(x)U(x)$. The diagonal factor $\Lambda(x)$ controls anisotropic scaling, while the orthogonal factor $U(x)$ is parameterized by a structured product of Givens rotations. This construction yields task-adaptive Dirichlet-regularized coordinates rather than eigenfunctions of a fixed supervision-independent operator. Experiments on synthetic, tabular, and image-based benchmarks study the effect of identity, diagonal, and rotation-scaling metrics, and compare the resulting coordinates with classical baselines and NeuralEF. The results support PIEFS as a compact supervised spectral representation method and identify optimization stability, \new{validation on explicit operator eigenproblems, and richer metric parameterizations as the main directions for future work}.
\end{abstract}

\section{Introduction}

\paragraph{Contributions.}
\begin{itemize}
    \item We formulate \textsc{piefs} as a supervised spectral-style representation method with sequential coordinate maps, empirical Gram orthogonality, cross-entropy on a linear readout, and a modified Dirichlet penalty.
    \item We study three metric settings inside the Dirichlet term: identity (\textsc{off}), volume-preserving diagonal scaling (\textsc{diag}), and diagonal scaling after a structured Givens rotation (\textsc{trotter}; apply order fixed as in Sec.~\ref{subsec:matrix_parametrization}).
    \item We evaluate the same training pipeline across five benchmark settings (Table~\ref{tab:main_results}) against RF, LR, PCA+LR, and NeuralEF~\citep{pmlr-v162-deng22b}, with visualizations of learned maps and training dynamics.
\end{itemize}

\label{sec:introduction}
Graph-based spectral methods build a Laplacian on finite samples and use its eigenmaps as geometry-aware features~\citep{4476091,10.1137/1.9781611972801.49}.
Their theory connects graph eigenfunctions to Laplace--Beltrami modes when data concentrate on a manifold~\citep{BELKIN20081289}, which explains strong performance in clustering and semi-supervised learning~\citep{NIPS2001_801272ee}.
Two bottlenecks motivate mesh-free alternatives: eigencomputation cost grows with dataset size~\citep{FORD201579}, and test-time evaluation at new points typically requires rebuilding the graph and its spectrum~\citep{BELKIN20081289}.
Neural surrogates for operator eigenproblems are an active line of work~\citep{Jin2020UnsupervisedNN,pmlr-v162-deng22b,fermionic_neural_network_states}.
Automatic differentiation makes it feasible to minimize Rayleigh-type objectives with PDE-style regularity while remaining discretization-free in the input domain~\citep{Feld_2019,LAGARIS19971}.

We study \emph{physics-informed eigenfunction features with learnable scaling} (\textsc{piefs}), implemented as learnable-metric Dirichlet coordinates: a sequence of scalar maps $(\phi_j)$ trained with a modified Dirichlet penalty that applies a data-dependent linear map $\mathbf{A}(\mathbf{x})$ to gradients~\citep{evans2022partial}, together with batch Gram orthogonality and cross-entropy on linear logits.
The outer schedule freezes earlier coordinates and updates one map at a time, yielding cheap inference once training finishes.
Section~\ref{sec:methodology} derives the losses and weighting; Section~\ref{sec:experimental_results} reports benchmarks and visualizations; Section~\ref{sec:discussion} discusses empirical findings, limitations, and future directions.

\section{Methodology}
\label{sec:methodology}
This section defines the \textsc{piefs} objective, the sequential training schedule, and the learnable metric used in the Dirichlet penalty.
We first derive the loss as a weighted sum of three terms, then the dynamic weights, then the learnable gradient metric $\mathbf{A}(\mathbf{x})$.
An optional graph-Laplacian warm-start could set auxiliary targets but is \emph{not} used in the experiments below.

\subsection{Loss Function}
\label{subseq:loss_function}
We motivate the coordinate construction with a Rayleigh-type variational picture: classical spectral problems minimize a Rayleigh quotient of Dirichlet form subject to orthogonality, whereas here we minimize a \emph{composite} objective that couples modified Dirichlet energy, a batch Gram surrogate for orthogonality, and cross-entropy on linear logits---this is not equivalent to solving a fixed self-adjoint eigenproblem on $L^2(p)$.
We consider a probability distribution with density $p(\mathbf{x})$.
This distribution defines the following inner product:
\begin{equation}
    \label{eq:bilinear_form_0}
    \langle f_1, f_2 \rangle_0 = \int f_1(\mathbf{x}) f_2(\mathbf{x}) p(\mathbf{x}) d\mathbf{x}
\end{equation}
Here $f_1$ and $f_2$ are functions with finite second moments.

The corresponding Dirichlet bilinear form is:
\begin{equation}
    \label{eq:bilinear_form_1_original}
    \langle f_1, f_2 \rangle_{\text{de}}
    = \int \sum_{a=1}^{d}
       \frac{\partial f_1(\mathbf{x})}{\partial x_a}
       \frac{\partial f_2(\mathbf{x})}{\partial x_a}
       \, p(\mathbf{x}) \,\mathrm{d}\mathbf{x}
\end{equation}
Here $d$ is the dimension of $\mathbf{x}$.

This product becomes a regular Dirichlet Energy for $f_1 = f_2$:
\begin{equation}
    \label{eq:dirichlet_energy_original}
    \begin{split}
    \langle f, f \rangle_{\text{de}}
    &= \int \biggl(
         \sum_{a=1}^{d}
         \frac{\partial f(\mathbf{x})}{\partial x_a}
         \frac{\partial f(\mathbf{x})}{\partial x_a}
       \biggr) p(\mathbf{x}) \,\mathrm{d}\mathbf{x} \\
    &= \int \lVert \nabla f(\mathbf{x}) \rVert^2\, p(\mathbf{x}) \,\mathrm{d}\mathbf{x}
    \end{split}
\end{equation}

We introduce a learnable matrix $\mathbf{A}(\mathbf{x})$ to modify the Dirichlet energy and the corresponding
bilinear form:

\begin{equation}
    \label{eq:bilinear_form_1}
    \langle f_1, f_2 \rangle_{\text{mde}}
    = \int \bigl(\mathbf{A}(\mathbf{x})\nabla f_1(\mathbf{x})\bigr)^{\!\top}
           \bigl(\mathbf{A}(\mathbf{x})\nabla f_2(\mathbf{x})\bigr)
           \, p(\mathbf{x}) \,\mathrm{d}\mathbf{x}
\end{equation}

For $f_1{=}f_2{=}f$, this gives the Modified Dirichlet Energy:
\begin{equation}
    \label{eq:dirichlet_energy}
    \begin{split}
    \langle f, f \rangle_{\text{mde}}
    &= \int \bigl\lVert \mathbf{A}(\mathbf{x})\nabla f(\mathbf{x}) \bigr\rVert^2
       \, p(\mathbf{x}) \,\mathrm{d}\mathbf{x}
    \end{split}
\end{equation}

The goal is to obtain $K$ coordinate maps $\phi_1(\mathbf{x}), \ldots, \phi_K(\mathbf{x})$ whose Dirichlet energy is small under~\eqref{eq:dirichlet_energy} while remaining approximately orthonormal and predictive under cross-entropy; block-coordinate descent on the composite loss approximates this Rayleigh-type trade-off rather than enforcing a single Rayleigh quotient.
Matrix $\mathbf{A}(\mathbf{x})$ is learned jointly with the basis so that a linear classifier atop $(\phi_1,\ldots,\phi_K)$ obtains high accuracy.

In other words, we have a loss function that corresponds to the MDE:
\begin{equation}
    \label{eq:loss_mde}
    \mathcal{L}_{\text{mde}} = \int \bigl\lVert \mathbf{A}(\mathbf{x}) \nabla \phi_k(\mathbf{x}) \bigr\rVert^2 \, p(\mathbf{x}) \,\mathrm{d}\mathbf{x}
\end{equation}
where $\phi_k$ is the currently active coordinate map at outer index $k$ during the sequential training schedule.
\new{There is a loss function that enforces the orthonormality constraint at the population level:
\begin{equation}
    \label{eq:loss_orth}
    \begin{split}
    \mathcal{L}_{\text{orth}}
    &= \big\| \mathbb{E}_{p}[\phi\phi^{\top}] - I_K \big\|_F^{2} \\
    &= \sum_{\alpha, \beta=1}^{K} \Big(\mathbb{E}_{p}\big[\phi_{\alpha}(\mathbf{x})\,\phi_{\beta}(\mathbf{x})\big] - \delta_{\alpha\beta}\Big)^{2},
    \end{split}
\end{equation}
where $\delta_{\alpha\beta}$ is the Kronecker symbol.}

\new{\begin{remark}[Monte-Carlo Gram estimator]
\label{remark:batch_gram}
The empirical Gram matrix $\hat C_k = \frac{1}{B}\Phi_{1:k}^{\top}\Phi_{1:k}$ is an unbiased estimator of $\mathbb{E}_{p}[\phi\phi^{\top}]$, so $\|\hat C_k - I_k\|_F^{2}$ minimised by SGD is a plug-in (bias-of-the-mean) estimator of~\eqref{eq:loss_orth}; the two coincide as $B\to\infty$, and the residual gap is $O(1/B)$ in expectation (variance of $\phi_\alpha\phi_\beta$ across the batch).
\end{remark}}

Features $\phi_1(\mathbf{x}), \ldots, \phi_K(\mathbf{x})$ are used to train the classifier:
\begin{equation}
    \label{eq:class_log_prob}
    \log\big(p(l|\mathbf{x})\big) \propto B_l + \sum_{\gamma=1}^{K} W_{l\gamma} \phi_{\gamma}(\mathbf{x})
\end{equation}
Here $B$ is the bias, $W$ stores the weights, $\log(p(l|\mathbf{x}))$ denotes the logits.
Weights and bias are learned from minimization of the cross-entropy loss function $\mathcal{L}_{\text{class}}$.
  
Finally, all three loss functions are combined together via weighting:
\begin{equation}
    \label{eq:loss_function}
    \begin{split}
    \mathcal{L} = \mathrm{sg}(w_{\text{orth}})\,\mathcal{L}_{\text{orth}}
               &+ \mathrm{sg}(w_{\text{class}})\,\mathcal{L}_{\text{class}} \\
               &+ \mathrm{sg}(w_{\text{mde}})\,\mathcal{L}_{\text{mde}}
    \end{split}
\end{equation}
Here $\mathrm{sg}(\cdot)$ denotes a stop-gradient operation: the adaptive weights are treated as constants during back-propagation.

Integrals over $p$ enter the loss functions as in Eq.~\eqref{eq:loss_mde} and Eq.~\eqref{eq:loss_orth}.
In practice, the density of the distribution is not known.
Therefore, Monte Carlo estimates replace these integrals in Eq.~\eqref{eq:loss_mde} and Eq.~\eqref{eq:loss_orth}.
Operationally, each integral is replaced by a batch average.

\subsection{Sequential Neural Coordinate Maps}
\label{subsec:seq_nn_ef}

We represent each coordinate map $\phi_j$ by a scalar neural ansatz with parameters $\theta_j$, chosen disjoint across $j$.
At outer index $k$ we optimize only $\theta_k$; for every $j<k$ the maps enter the Gram and classification losses as \emph{fixed} nonlinearities, so gradients with respect to $\theta_j$ vanish identically and the empirical Gram $\hat C_k$ is consistent with a block-coordinate descent on $(\theta_1,\ldots,\theta_K)$.
The Dirichlet penalty involves $\nabla_{\mathbf{x}}\phi_k$ alone.
The classifier logits append zeros in coordinates that have not yet been activated; that is, coordinates with index $j>k$ contribute zero to the classification loss throughout their development phase, which is equivalent to evaluating the affine readout on the leading $k$-dimensional coordinate subspace until the schedule completes.

We do not identify $(\phi_j)$ with eigenfunctions of a prescribed self-adjoint operator on $L^2(p)$; rather, they form orthogonal, Dirichlet-regularized coordinates biased by cross-entropy, analogously to learned spectral features~\citep{pmlr-v162-deng22b}.

The BasisSet learns $(\phi_j)$ sequentially (one active outer index at a time) while MetricNet supplies $\mathbf{A}(\mathbf{x})=\boldsymbol{\Lambda}(\mathbf{x})\mathbf{U}(\mathbf{x})$; both couple through~\eqref{eq:dirichlet_energy}--\eqref{eq:loss_function}.

Algorithm~\ref{alg:piefs-seq} lists the sequential optimization schedule.
At outer index~$k$, the frozen stack $\boldsymbol{\Phi}_{1{:}k-1}$ and the trainable map $\phi_k$ jointly feed $\mathcal{L}_{\mathrm{orth}}$ and $\mathcal{L}_{\mathrm{class}}$, while $\mathcal{L}_{\mathrm{mde}}$ couples $\mathbf{A}(\mathbf{x})$ to $\nabla_{\mathbf{x}}\phi_k$ (see~\eqref{eq:dirichlet_energy}); only $\theta_k$ receives gradients (all $\theta_j$ with $j \neq k$ are detached via stop-gradient).
The losses aggregate as in~\eqref{eq:loss_function} with weights~\eqref{eq:loss_function_weight}.
After the schedule completes, inference uses the affine readout on the leading coordinates~\eqref{eq:class_log_prob}.

\begin{algorithm}[tb]
 \caption{Sequential block-coordinate descent for \textsc{piefs}.}
 \label{alg:piefs-seq}
 \begin{algorithmic}[1]
 \FOR{$k = 1$ to $K$}
   \FOR{each inner iteration until the budget is exhausted}
    \STATE Fix $\theta_j$ for all $j<k$.
    \STATE Evaluate $\Phi_{1:k}$ by applying $\phi_j(\cdot;\theta_j)$ with fixed $\theta_j$ for $j<k$ and the current iterate $\theta_k$ for $j=k$.
    \STATE Accumulate $\|\hat C_k-I_k\|_F^{2}$, the Dirichlet contribution $\|\mathbf{A}\nabla\phi_k\|^2$, and cross-entropy computed on logits whose coordinates with index $j>k$ are fixed to zero.
    \STATE Refresh $(w_{\mathrm{orth}},w_{\mathrm{class}},w_{\mathrm{mde}})$ via~\eqref{eq:loss_function_weight} with $g_k=\|\hat C_k-I_k\|_F^{2}$, treating these weights as constants in the differentiation of $\mathcal{L}$.
   \ENDFOR
 \ENDFOR
 \end{algorithmic}
\end{algorithm}

\subsection{Loss Function Weighting}
\label{subsec:loss_function_weigting}
We use the following hierarchy of loss functions:
\begin{itemize}
    \item orthonormality
    \item cross-entropy for classification
    \item MDE loss function
\end{itemize}
The objective is to learn representative features first and train the classifier with them.
The final stage of MDE minimization provides regularization and improves the expressive power of constructed features.

Throughout all reported experiments, we fix two temperature scales: $T_{\mathrm{orth}}{=}0.1$ and $T_{\mathrm{class}}{=}0.5$.
These values were selected via informal grid search on the MNIST validation set and held constant across all datasets;
a formal hyperparameter ablation remains an open direction for future work.

The practical way to achieve the hierarchy of concern is dynamic weighting of loss functions that is updated
each iteration of loss function minimization:
Let $g_k = \|\hat C_k-I_k\|_F^{2}$.
\begin{equation}
    \label{eq:loss_function_weight}
    \left\{\begin{aligned}
        w_{\text{orth}}  &\propto 1, \\
        w_{\text{class}} &\propto \exp\!\bigl(- g_k / T_{\text{orth}}\bigr), \\
        w_{\text{mde}}   &\propto \exp\!\Bigl(- \max\bigl( g_k / T_{\text{orth}},\; \mathcal{L}_{\text{class}} / T_{\text{class}}\bigr)\Bigr).
    \end{aligned}\right.
\end{equation}
Here $T_{\text{orth}}$ and $T_{\text{class}}$ are temperature-style scales fixed \emph{a priori} for the reported runs ($T_{\text{class}}{=}0.5$); an optional graph-based warm-start could in principle set them from held-out logistic loss~\cite{4476091}, but we do not use that stage here.

The weighting in~\eqref{eq:loss_function_weight} employs exponential dampening when loss values are large.
The exponential form was chosen empirically; alternative functional forms such as $1/(1+g_k)$
or polynomial dampening were tested informally and found less robust in early experiments, though
systematic ablation of this choice remains an open question for future work.
Equation~\eqref{eq:loss_function_weight} implements the desired priority order: when $g_k$ is large, the classification and MDE terms are suppressed; the MDE term becomes active only after both Gram and classification residuals are moderate.

\subsection{Matrix Parametrization and \textsc{trotter}-Fixed Rotations}
\label{subsec:matrix_parametrization}
Equation~\eqref{eq:dirichlet_energy} is evaluated using the gradient of the active map $\phi_k$ only~(Sec.~\ref{subsec:seq_nn_ef}).
We summarize the parametrizations for $\mathbf{A}(\mathbf{x})$ that appear in Table~\ref{tab:main_results}.

At the continuum level one may factor
\begin{equation}
    \label{eq:amat_factorisation}
    \mathbf{A}(\mathbf{x}) = \mathbf{\Lambda} (\mathbf{x}) \mathbf{U}(\mathbf{x}),
\end{equation}
with diagonal $\boldsymbol{\Lambda}$ and orthogonal $\mathbf{U}$.
Throughout training, $\mathbf{A}(\mathbf{x})$ is fit jointly with the coordinate nets and the classifier through the computation graph that evaluates~\eqref{eq:loss_function}; cross-entropy and the adaptive weights in~\eqref{eq:loss_function_weight} therefore couple $\mathbf{A}$ to label information, so it should not be read as an unsupervised surrogate for a fixed physics-only differential operator.
\textbf{Diagonal scaling.} The log-eigenvalues follow a zero-mean MLP head so that
\begin{equation}
    \label{eq:diagonal_constraint}
    \sum_{i=1}^{d} \log \lambda_i(\mathbf{x}) = 0
    \quad\Rightarrow\quad
    \det \boldsymbol{\Lambda}(\mathbf{x}) = 1.
\end{equation}

A fully expressive $\mathrm{SO}(d)$ parametrization would require substantially more degrees of freedom, for example through dense skew-symmetric generators or matrix exponentials.
Such variants are computationally expensive at the ambient dimensions considered here.
We therefore use a structured Givens product as a stable low-parameter baseline.

\paragraph{Trained metric variants.}
\textbf{Metric disabled (\textsc{off}).}
We set $\mathbf{A}=\mathbf{I}$ identically while still learning orthogonal features; Euclidean Dirichlet energies arise as the special case of~\eqref{eq:dirichlet_energy}.

\textbf{Diagonal (\textsc{diag}).} $\mathbf{A}(\mathbf{x})=\boldsymbol{\Lambda}(\mathbf{x})$ with coordinates $\lambda_i(\mathbf{x})=\exp(z_i(\mathbf{x})-\bar z(\mathbf{x}))$ under~\eqref{eq:diagonal_constraint}.

\textbf{Trotter rotation (\textsc{trotter}).} We parameterize $\mathbf{U}(\mathbf{x})$ as a product of $d{-}1$ Givens rotations,
\begin{equation}
    \mathbf{U}(\omega) = R_{d-2}(\omega_{d-2})\cdots R_1(\omega_1)\,R_0(\omega_0),
\end{equation}
with angles $\omega_i(\mathbf{x})=\pi\tanh\bigl(\mathrm{MLP}(\mathbf{x})_i\bigr)$.
Each plane rotation $R_j$ lies in $\mathrm{SO}(d)$, so the product is orthogonal with determinant one; this fixed chain is a \emph{structured} subset of orthogonal maps and does not parameterize all of $\mathrm{SO}(d)$ (which has dimension $d(d{-}1)/2$), but it remains an expressive, low-parameter family for high ambient $d$.
Bounding angles via $\omega_i = \pi\tanh(\cdot)$ keeps them in $[-\pi,\pi]$, improving numerical stability
and preventing unbounded extrapolation in the angle parameterization.
The learned metric acts on tangent vectors $\mathbf{v}$---in practice $\mathbf{v}=\nabla\phi_k$---by rotating first and rescaling afterwards,
    \begin{equation}
    \label{eq:amat_apply_order}
    \mathbf{A}(\mathbf{x})\mathbf{v} = \boldsymbol{\Lambda}(\mathbf{x})\big(\mathbf{U}(\mathbf{x})\mathbf{v}\big),
    \end{equation}
which we adopt throughout this work.
\textbf{Gradient flow.}
Equation~\eqref{eq:amat_apply_order} exposes both factors to stochastic gradients via $\|\mathbf{A}(\mathbf{x})\mathbf{v}\|^2=\sum_i \lambda_i^2\big((\mathbf{U}\mathbf{v})_i\big)^2$.
If $\boldsymbol{\Lambda}$ is applied before $\mathbf{U}$, the Euclidean energy $\|\mathbf{A}\mathbf{v}\|^2$ collapses to a diagonal-only functional and no longer differentiates through the rotational factor; all experiments below therefore adopt~\eqref{eq:amat_apply_order}.
\textbf{Network architecture.}
All basis functions and metric networks use three fully-connected hidden layers of width 64 with ReLU activations.
Basis functions $\phi_k$ output a single scalar; metric networks output $K$ values (for the diagonal variant)
or $(d-1)$ angle values (for the Trotter variant, one per adjacent pair rotation).

\textbf{Numerical contraction.}
Adjacent plane rotations are fused into even--odd sweeps with depth bounded independently of the nominal Givens count, following standard practice for compact differentiable orthogonal maps.

PIEFS-\textsc{trotter} on Two Moons, Circles, and HTRU2 aggregates five independent optimization trajectories, each with $60{,}000$ gradient steps, under~\eqref{eq:amat_apply_order}.
The same five-seed protocol applies to MNIST multiclass \textsc{trotter} at $60{,}000$ steps, whereas CIFAR-10 ResNet-18 embeddings use $120{,}000$ total steps per seed when \textsc{trotter} is enabled.

\subsection{Pretraining}
\label{subsec:pretraining}
Optional Graph-Laplacian pretraining could fit eigenmaps on a subsample and calibrate $T_{\mathrm{class}}$ from a linear probe~\cite{4476091}. \textbf{All experiments below} disable this stage, fix $T_{\mathrm{class}}=0.5$, and train \textsc{piefs} from scratch without graph-based warm starts. \new{A small five-seed ablation on HTRU2 with the warm start enabled is reported in App.~\ref{app:gl_ablation}.}

\section{Experimental Results}
\label{sec:experimental_results}

\paragraph{Benchmark overview.}
We compare \textsc{piefs} to classical baselines and NeuralEF~\citep{pmlr-v162-deng22b} on five benchmark settings summarized in Table~\ref{tab:main_results}; splits, coordinate counts~$K$, validation vs.\ test reporting, and baseline choices are detailed in Sec.~\ref{subsec:datasets}.

\paragraph{Default \textsc{piefs} protocol.}
Unless otherwise stated, all \textsc{piefs} runs use the same training procedure: sequential coordinate-map optimization, empirical Gram orthogonality, cross-entropy on a linear readout, dynamic loss weights, no graph-Laplacian pretraining, and no input augmentation. The columns \textbf{off}, \textbf{diag}, and \textbf{trotter} differ only in the metric $\mathbf{A}(\mathbf{x})$ used inside the Dirichlet penalty.

\begin{table}[t]
  \centering
  \footnotesize
  \setlength{\tabcolsep}{5pt}
  \renewcommand{\arraystretch}{1.08}
  \caption{Training components used in the reported experiments.}
  \label{tab:training_components}
  \begin{tabular}{@{}p{0.34\linewidth} p{0.60\linewidth}@{}}
    \toprule
    \textbf{Component} & \textbf{Role} \\
    \midrule
    Sequential coordinate training & Trains one map $\phi_k$ at a time \\
    Gram orthogonality & Empirical penalty encouraging approximate coordinate independence \\
    Cross-entropy readout & Linear classifier on coordinates; makes features task-adaptive \\
    Modified Dirichlet penalty & Encourages smooth coordinates via weighted input gradients \\
    Graph-Laplacian pretraining & Optional warm start (not used in reported runs) \\
    Metric OFF & Dirichlet uses $\mathbf{A}(\mathbf{x})=\mathbf{I}$ \\
    Metric DIAG & Dirichlet uses diagonal scaling $\mathbf{A}(\mathbf{x})=\boldsymbol{\Lambda}(\mathbf{x})$ \\
    Metric TROTTER & Dirichlet uses $\mathbf{A}(\mathbf{x})=\boldsymbol{\Lambda}(\mathbf{x})\mathbf{U}(\mathbf{x})$ (apply order as in Sec.~\ref{subsec:matrix_parametrization}) \\
    \bottomrule
  \end{tabular}
\end{table}

\begin{table*}[t]
  \caption{Accuracy (\%, mean$\pm$std). \textsc{piefs} columns report test-split accuracy over five seeds. Classical baselines use the splits described in Sec.~\ref{subsec:datasets}: validation split for Two Moons, Circles, and HTRU2, and official test split for MNIST and CIFAR-10 (R18 emb.). NeuralEF\textsuperscript{$\ast$} reports five-seed test accuracy: for MNIST and CIFAR-10 (R18 emb.) from our rerun of the public codebase with $K{=}16$ linear probe, and for Two Moons, Circles, and HTRU2 (no public-codebase numbers available) from an RBF-kernel NeuralEF at the dataset-matched $K$. Bold marks the best \textsc{piefs} variant in each row. Note that NeuralEF\textsuperscript{$\ast$} is \emph{unsupervised} whereas the \textsc{piefs} columns here are \emph{supervised}; a like-for-like unsupervised comparison is reported in App.~\ref{app:unsup_ablation} (Table~\ref{tab:unsup_ablation}).}
  \label{tab:main_results}
  \centering
  \footnotesize
  \setlength{\tabcolsep}{3.5pt}
  \resizebox{\textwidth}{!}{%
  \begin{tabular}{lccccccc}
    \toprule
    \textbf{Dataset} & \textbf{RF} & \textbf{LR} & \textbf{PCA+LR} & \textbf{NeuralEF}\textsuperscript{$\ast$} & \textbf{PIEFS-off} & \textbf{PIEFS-diag} & \textbf{PIEFS-trotter} \\
    \midrule
    Two Moons  & \new{$99.89{\pm}0.04$} & \new{$88.15{\pm}0.22$} & \new{$88.15{\pm}0.22$} & \new{$88.23{\pm}0.31$}  & \new{$\mathbf{99.87}{\pm}0.00$} & \new{$99.84{\pm}0.08$} & \new{$99.88{\pm}0.03$} \\
    Circles    & \new{$99.96{\pm}0.04$} & \new{$50.67{\pm}5.12$} & \new{$50.67{\pm}5.12$} & \new{$100.00{\pm}0.00$}  & \new{$99.99{\pm}0.03$} & \new{$99.97{\pm}0.06$} & \new{$\mathbf{100.00}{\pm}0.00$} \\
    HTRU2      & \new{$98.16{\pm}0.28$} & \new{$97.88{\pm}0.27$} & \new{$97.88{\pm}0.28$} & \new{$97.47{\pm}0.34$}  & \new{$97.90{\pm}0.15$}           & \new{$\mathbf{97.97}{\pm}0.06$} & \new{$97.92{\pm}0.10$} \\
    \midrule
    MNIST      & \new{$97.07{\pm}0.03$} & \new{$92.20{\pm}0.00$} & \new{$85.94{\pm}0.00$} & $82.52{\pm}0.29$ & \new{$\mathbf{95.41}{\pm}0.32$}  & \new{$94.91{\pm}0.28$} & \new{$94.94{\pm}0.20$} \\
    CIFAR-10 (R18 emb.) & \new{$82.65{\pm}0.11$} & \new{$86.82{\pm}0.00$} & \new{$78.11{\pm}0.00$} & $76.05{\pm}0.55$ & $\mathbf{85.50}{\pm}0.53$  & $84.98{\pm}0.33$ & $85.22{\pm}0.39$ \\
    \bottomrule
  \end{tabular}%
  }%
\end{table*}

\begin{figure*}[t]
  \centering
  \includegraphics[width=\paperfigwidthGeom,height=\figHgeom,keepaspectratio]{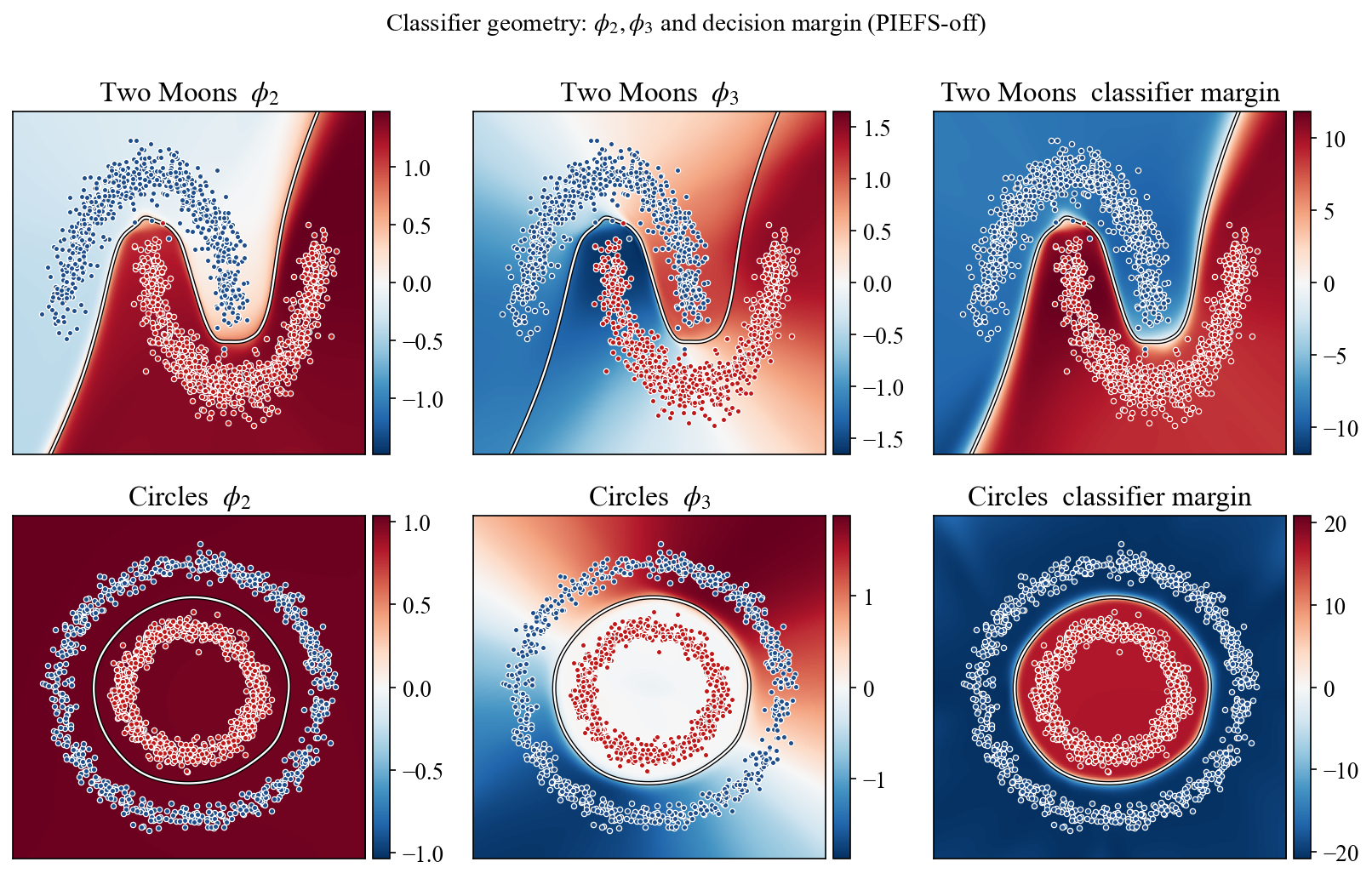}
  \caption{Classifier geometry induced by \textsc{piefs} coordinates. Columns 1--2 show $\phi_2$ and $\phi_3$ on Two Moons and Circles (PIEFS-off, one seed). Column 3 shows the signed classifier margin (linear-readout logit) pulled back to the input plane $(x_1,x_2)$. Figure~\ref{fig:eigenfunctions} shows the corresponding individual coordinate heatmaps for the same replicate.}
  \label{fig:classifier_geometry}
\end{figure*}

\subsection{Experimental Setup}
\label{subsec:datasets}

The five benchmark settings in Table~\ref{tab:main_results} span synthetic binary tasks, tabular data, multiclass images, and CIFAR-10 ResNet-18 embeddings.
The following paragraphs specify datasets, splits, coordinate counts~$K$, and what is reported as validation vs.\ test.

\textbf{Binary datasets.}
Synthetic benchmarks offer controlled evaluation and interpretability; accordingly, \emph{Two Moons} and \emph{Circles} are planar synthetic tasks with $n{=}10{,}000$ samples (Gaussian noise $\sigma{=}0.1$ for Two Moons and $\sigma{=}0.05$ for Circles) and a \(70\%/15\%/15\%\) split between training, validation, and test data~\citep{scikit-learn}.
\emph{HTRU2}~\citep{10.1093/mnras/stw656,Lyon2016} provides eight tabular features for pulsar candidate classification
on $17{,}898$ samples with the same split and z-scored inputs.

\textbf{Multiclass and embedding benchmarks.}
\emph{MNIST}~\citep{lecun-mnisthandwrittendigit-2010} is used at native resolution with $784$ input dimensions; we use the standard 60k training and 10k test split, further subdividing the test set into 5k validation and 5k held-out evaluation; we set $K{=}16$.
\emph{CIFAR-10}~\citep{Krizhevsky2009LearningML} is represented by $512$-dimensional penultimate activations of a pretrained ResNet-18 using the standard 45k training, 5k validation, and 10k test partition; we set $K{=}16$.
Classical baselines use standardisation for LR/PCA+LR and raw $[0,1]$ inputs for RF.

\textbf{Protocols.} Unless noted otherwise, Table~\ref{tab:main_results} reports \textsc{piefs} \textbf{test}-split accuracy with mean and standard deviation over five random seeds.
The NeuralEF column for MNIST and CIFAR-10 ResNet-18 embeddings reports \textbf{test} accuracy (mean $\pm$ std, five seeds, $K{=}16$ linear probe) from our rerun of the public codebase~\citep{pmlr-v162-deng22b}; Deng et al.\ additionally report \textbf{84.98}\% MNIST test for their strongest CNN-GP configuration (Table~1), which is not identical to our linear-probe setting.
Basis networks ($\phi_k$) use three hidden layers of width $64$; metric networks ($\boldsymbol{\Lambda}$, $\mathbf{U}$) use two hidden layers of width $64$.

\paragraph{Classical baselines.}
Random forests (200 trees) and multinomial logistic regression on raw or standardised features; \textbf{PCA+LR} applies PCA with $16$ components followed by the same logistic regression~\citep{scikit-learn}. Validation split for low-dimensional tasks, official test for MNIST and CIFAR-10 embeddings.

\paragraph{NeuralEF baseline.}
\textbf{NeuralEF}~\citep{pmlr-v162-deng22b} is an unsupervised spectral baseline with a linear probe; we report five-seed test numbers for MNIST and CIFAR-10 embeddings from our rerun of the public codebase under the same $K{=}16$ linear-probe protocol. \new{For the low-dimensional and tabular datasets (Two Moons, Circles, HTRU2), where no public-codebase configuration is provided, we use an RBF-kernel NeuralEF implementation at the dataset-matched $K$ with the same linear-probe evaluation.}

\subsection{Main Results}

The three \textsc{piefs} columns in Table~\ref{tab:main_results} are \textbf{off} ($\mathbf{A}{=}\mathbf{I}$), \textbf{diag} ($\mathbf{A}{=}\boldsymbol{\Lambda}$), and \textbf{trotter} ($\mathbf{A}{=}\boldsymbol{\Lambda}\mathbf{U}$ with Sec.~\ref{subsec:matrix_parametrization} apply order).
Five-seed \textsc{trotter} on Two Moons, Circles, HTRU2, and MNIST follows the same protocol as the other rows; CIFAR-10 (R18 emb.) \textsc{trotter} is likewise aggregated over five independent seeds with the step budget stated above.

\paragraph{Result highlights.}
On the \emph{linearly separable} HTRU2 dataset, both LR and PIEFS-off match RF,
confirming the method does not degrade when the raw features are already informative.
On \emph{Two Moons}, where LR achieves only 88.2\% (the linear boundary cannot separate the two crescents),
\new{PIEFS-off reaches 99.87\%}, demonstrating nonlinear separability beyond a linear boundary on raw inputs.
On \emph{MNIST}, PIEFS-off reaches \(95.41\%\) \textbf{test} accuracy at $K{=}16$.
NeuralEF~\citep{pmlr-v162-deng22b} uses a linear probe without labels in eigenfeature training; Deng et al.\ report \textbf{84.98}\% MNIST test for their CNN-GP setup vs.\ \textbf{82.52$\pm$0.29}\% in our $K{=}16$ rerun (Table~\ref{tab:main_results}); neither matches PIEFS \new{test} at $K{=}16$.
On CIFAR-10 embeddings our NeuralEF rerun reaches \textbf{76.05$\pm$0.55}\% test vs.\ \textbf{85.50$\pm$0.53}\% PIEFS-off test.
\new{On \emph{Circles} (well-separated concentric rings), all three variants reach near-perfect test accuracy ($\geq 99.9\%$).}
On CIFAR-10 embeddings, PIEFS-diag trails PIEFS-off slightly, underscoring that extra scaling does not always help high-level features.

\subsection{Qualitative Geometry}
\label{subsec:qualitative_geometry}

\begin{figure*}[!t]
  \centering
  \includegraphics[width=\paperfigwidth,height=\figHbig,keepaspectratio]{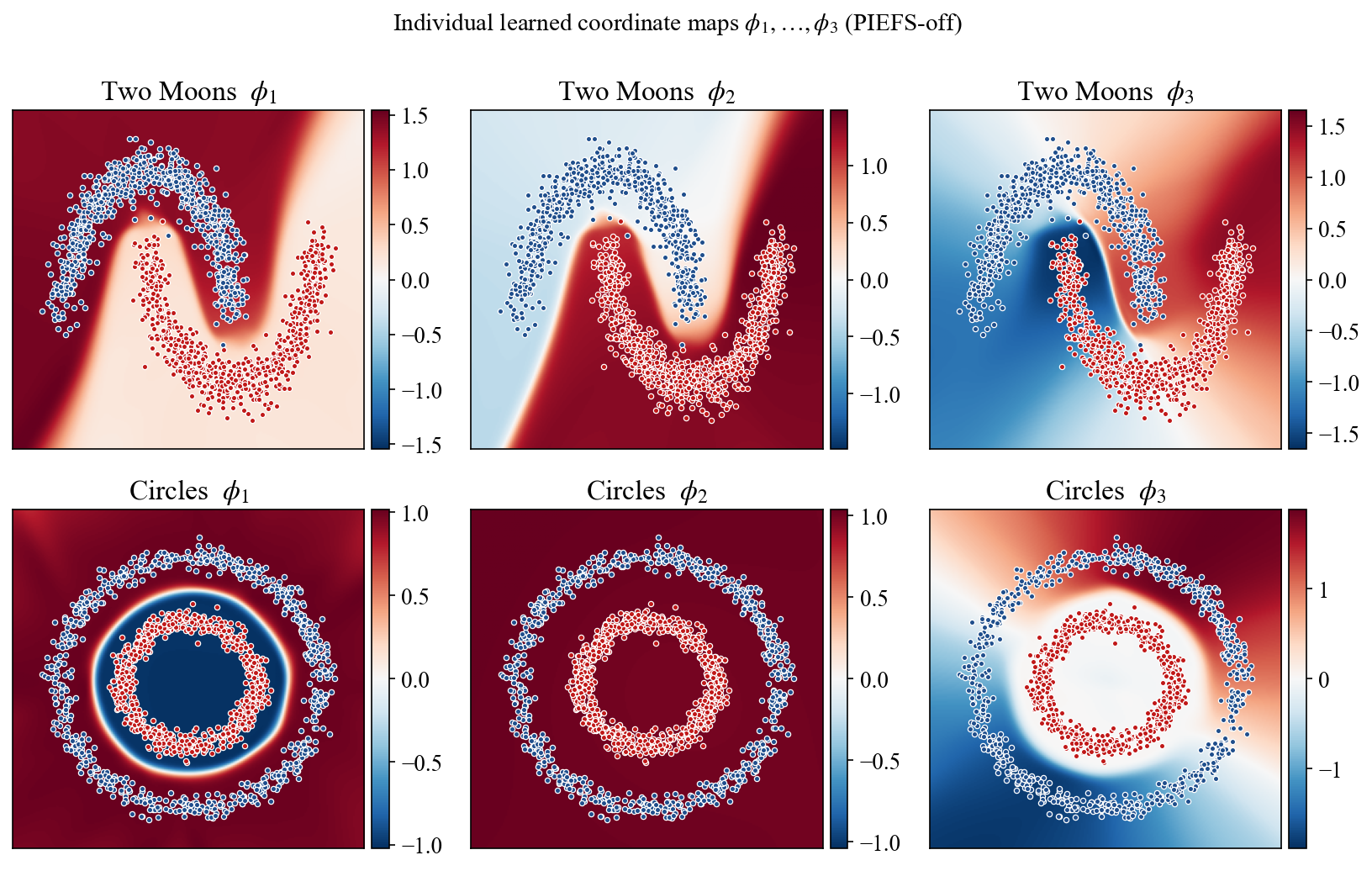}
  \caption{Leading learned coordinate maps $\phi_1,\phi_2,\phi_3$ (spectral view). Each panel shows one coordinate map $\phi_j(x_1,x_2)$ as a heatmap; dots are validation points colored by class. Unlike Fig.~\ref{fig:classifier_geometry}, no decision-boundary contour is overlaid: these are the raw learned coordinates, showing the dominant mode $\phi_1$ (absent from Fig.~\ref{fig:classifier_geometry}) and the progression to higher modes. PIEFS-off, same representative replicate as in Fig.~\ref{fig:classifier_geometry}.}
  \label{fig:eigenfunctions}
\end{figure*}

Figure~\ref{fig:classifier_geometry} visualizes how the learned coordinates induce nonlinear classifier margins in input space (Two Moons and Circles; PIEFS-off, one seed).
Figure~\ref{fig:eigenfunctions} shows the leading three coordinate maps $\phi_1,\phi_2,\phi_3$ for the same replicate without the decision-boundary overlay, highlighting the spectral progression from the dominant mode~$\phi_1$.

\subsection{Training Dynamics and Orthogonality}
\label{subsec:training_dynamics}

\begin{figure*}[!t]
  \centering
  \begin{subfigure}[b]{0.46\textwidth}
    \centering
    \includegraphics[width=\linewidth]{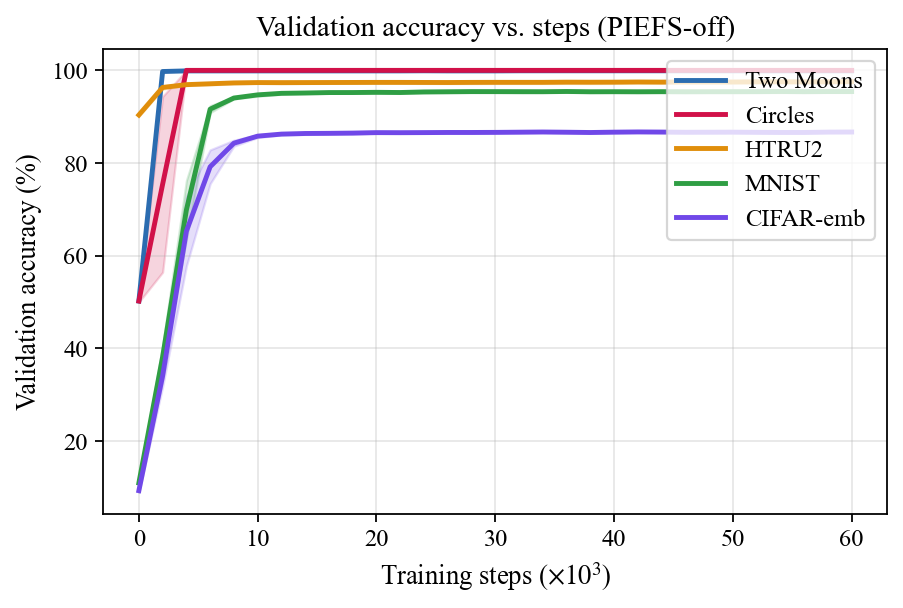}
    \caption{\new{Validation accuracy vs.\ training steps for PIEFS-off across all five datasets (mean $\pm$ std over seeds).}}
    \label{fig:training}
  \end{subfigure}\hfill
  \begin{subfigure}[b]{0.46\textwidth}
    \centering
    \includegraphics[width=\linewidth]{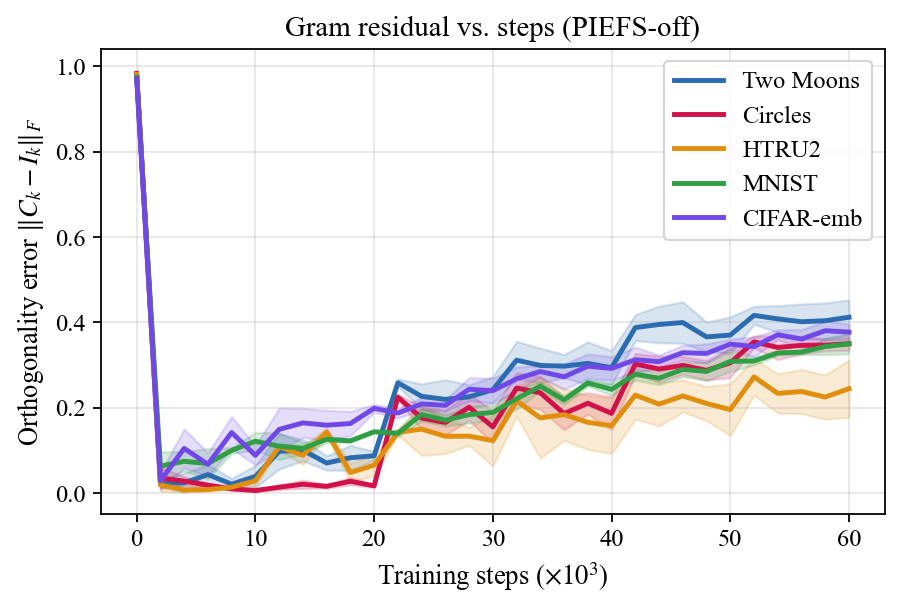}
    \caption{\new{Orthogonality error (Frobenius norm $\|C_k-I_k\|_F$ of the batch Gram matrix vs.\ identity) for PIEFS-off across all five datasets (mean $\pm$ std over seeds). Non-monotone curves reflect dynamic loss weighting~\eqref{eq:loss_function_weight}.}}
    \label{fig:gram_convergence}
  \end{subfigure}
  \caption{Training dynamics and orthogonality diagnostics \new{(PIEFS-off, all five datasets)}. (a)~Validation accuracy over training steps. (b)~Batch Gram residual $\|C_k-I_k\|_F$. Curves show mean $\pm$ std over seeds.}
  \label{fig:training_diagnostics}
\end{figure*}

Figure~\ref{fig:training_diagnostics} summarizes training dynamics and orthogonality diagnostics.
Panel~\subref{fig:training} shows validation accuracy over training steps for PIEFS-off on all five datasets (mean $\pm$ std over seeds).
\new{All datasets converge within the first 15{,}000 steps and remain stable thereafter.}
\new{The leading coordinates carry most of the discriminative signal: on MNIST the first four maps (step $15{,}000$, $4{\times}3750$) already reach $94.5\%$ validation accuracy, and the remaining twelve add only ${\sim}0.5$ percentage points, consistent with a spectral ordering in which early modes dominate.}
Panel~\subref{fig:gram_convergence} tracks the batch Gram residual $\|C_k-I_k\|_F$ for PIEFS-off at $K{=}16$; because the outer schedule reweights Gram, classification, and MDE terms, the curve need not be monotone.

\begin{figure}[!t]
  \centering
  \includegraphics[width=0.72\linewidth,keepaspectratio]{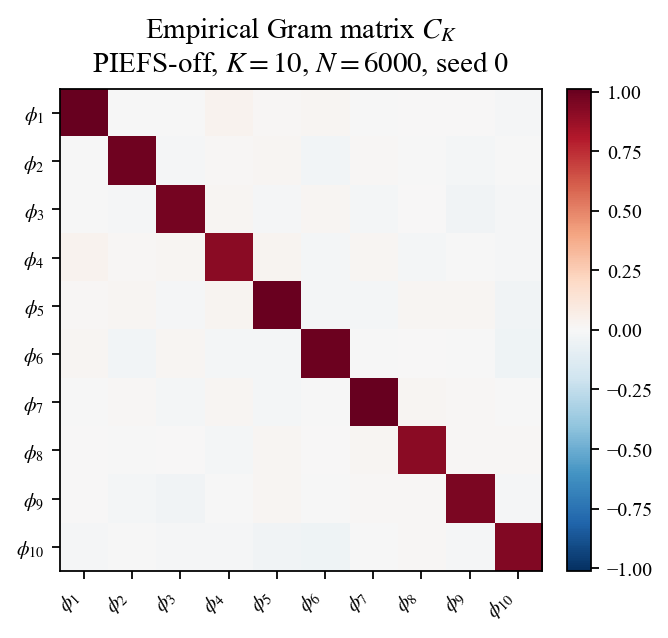}
  \caption{Empirical Gram matrix $C_K$ on MNIST validation. Off-diagonal entries are near zero and diagonal entries near one, confirming approximate orthonormality. PIEFS-off, $K{=}10$, seed~0.}
  \label{fig:gram_ck}
\end{figure}

\begin{figure}[!t]
  \centering
  \includegraphics[width=0.72\linewidth,keepaspectratio]{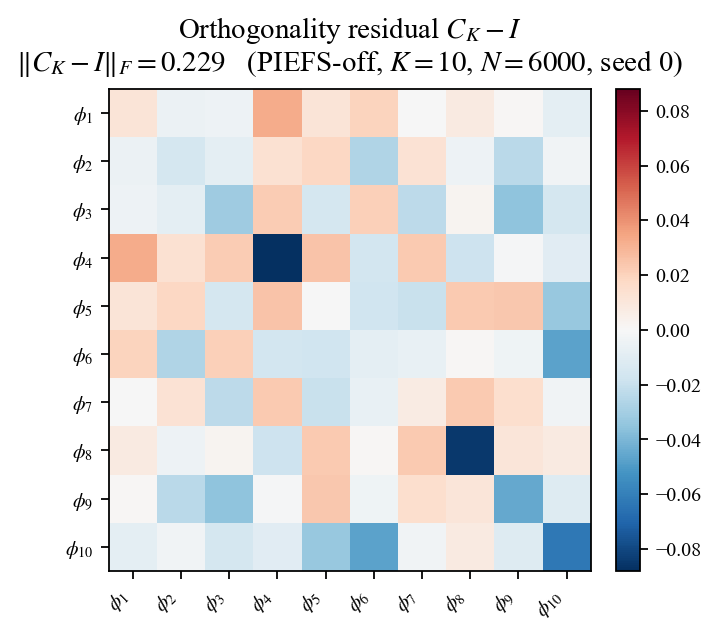}
  \caption{Orthogonality residual $C_K-I$ for the same replicate as Fig.~\ref{fig:gram_ck}. The largest deviations are ${\approx}0.09$; $\|C_K-I\|_F{=}0.23$, consistent with finite-batch estimation (Remark~\ref{remark:batch_gram}).}
  \label{fig:gram_residual}
\end{figure}

Figures~\ref{fig:gram_ck}--\ref{fig:gram_residual} show the empirical Gram matrix $C_K$ and its residual $C_K-I$ computed on the MNIST validation set after training (PIEFS-off, $K{=}10$, seed~0).
The matrix is close to the identity; the Frobenius deviation $\|C_K{-}I\|_F{=}0.23$ indicates approximate rather than exact orthogonality, which is expected under finite-batch estimation (Remark~\ref{remark:batch_gram}) and remains well-behaved in practice.

Gradients use automatic differentiation~\cite{NEURIPS2019_9015}; informal wall-clock remarks appear in App.~\ref{app:runtime_notes}.

The adaptive weights in~\eqref{eq:loss_function_weight} prioritize Gram and cross-entropy early, then emphasize MDE once residuals shrink, reducing conflict between oscillatory gradients and classification.

\section{Discussion and Limitations}
\label{sec:discussion}

We present \textsc{piefs}, a neural feature-learning method based on a modified Dirichlet-energy objective.
The method learns a sequence of scalar coordinate maps under an empirical Gram-orthogonality constraint, while a supervised cross-entropy term biases these coordinates toward the downstream classification task.
The resulting features should therefore be interpreted as task-adaptive, Dirichlet-regularized spectral coordinates rather than eigenfunctions of a fixed, supervision-independent differential operator.

The main empirical observation is that the learned coordinates provide useful nonlinear representations on several benchmarks.
On Two Moons, \textsc{piefs} separates a geometry that is not captured by a linear classifier on the raw input.
On MNIST and CIFAR-10 ResNet-18 embeddings, the learned coordinates also give competitive test accuracy with a compact feature dimension.
At the same time, the results do not uniformly dominate classical baselines: random forests and logistic regression remain strong on HTRU2.
These cases indicate that the method is effective across several geometries while not uniformly dominating strong classical baselines.

The learnable metric $\mathbf{A}(\mathbf{x})$ provides a practical way to modify the Dirichlet penalty.
The diagonal variant changes the relative scaling of gradient directions, while the \textsc{trotter} variant applies a product of Givens rotations followed by diagonal scaling.
This parameterization keeps the rotational factor orthogonal by construction and avoids runtime QR or Gram--Schmidt orthogonalization.
However, the current sparse Givens chain is not a full parameterization of $\mathrm{SO}(d)$, so its expressivity is limited.
Richer orthogonal factors (e.g.\ dense skew-symmetric generators or matrix exponentials) would better approximate $\mathrm{SO}(d)$ but are costly at the ambient dimensions~$d$ used here; the structured Givens product is our stable baseline.
Learning intermediate or hybrid rotation families remains an important open question.

A central limitation is the mathematical status of the learned coordinates.
Since both the basis functions and, in some variants, the metric are trained with label supervision, the method does not recover the spectrum of a prescribed self-adjoint operator.
Instead, it learns coordinates that balance three objectives: empirical orthogonality, classification accuracy, and Dirichlet-type smoothness.
This distinction is important for interpreting the method: \textsc{piefs} is closer to supervised spectral representation learning than to a classical eigenvalue solver.

Another limitation is the use of finite-batch Gram penalties.
The constraint $\widehat{C}_k \approx I_k$ is only a Monte Carlo approximation to global $L^2(p)$-orthogonality, and the residual may fluctuate during training as the classification and energy terms change the active coordinate.
The dynamic weighting scheme mitigates this interaction but does not provide a monotone guarantee.
This is visible in the Gram-residual traces (Fig.~\ref{fig:gram_convergence}), where the constraint can improve early and later deviate as other objectives become active.

The comparison with NeuralEF should also be interpreted with care.
NeuralEF is an unsupervised spectral-feature method, whereas \textsc{piefs} uses label information through the cross-entropy term.
The reported NeuralEF number is therefore a reference point for compact spectral representations, not a like-for-like comparison.
\new{An unsupervised ablation (App.~\ref{app:unsup_ablation}) repeats the pipeline with $w_{\mathrm{class}}{=}0$ for an unsupervised reference.}

The physics-informed component of \textsc{piefs} is the Modified Dirichlet Energy itself.
Dirichlet energies arise in variational formulations of diffusion-type operators, elliptic PDEs, and spectral problems; here, the geometry of this energy is made data-adaptive through the learned metric $A(x)$.
The experiments therefore evaluate a learned MDE-based representation method rather than a classical solver for a prescribed operator with known spectrum.

In summary, \textsc{piefs} provides a compact inductive feature map with a physics-inspired smoothness bias and inexpensive inference after training.
The method is promising as a supervised spectral representation learner, but its current form should not be presented as a general-purpose neural eigensolver.
The main directions for improvement are better optimization stability, richer metric parameterizations---especially flexible $\mathrm{SO}(d)$-valued factors beyond the implemented Givens chain---and validation on explicit PDE/operator eigenvalue problems.

\FloatBarrier

\section*{Acknowledgements}
The authors thank the HSE University supercomputer cluster (cHARISMa)~\citep{kostenetskiy2021charisma} for providing the computational resources used in this work.

Alexander Tarakanov gratefully acknowledges Nikolay Anokhin and Alexander D'yakonov of AI VK for fruitful discussions, valuable advice, and their generous support throughout the development of this work.

\bibliography{bibliobase}
\bibliographystyle{icml2026}

\clearpage
\appendix
\raggedbottom

\section{Extended Protocols and Notes}
\label{app:extended}

\subsection{Implementation and Runtime}
\label{app:runtime_notes}
Optimization uses mini-batch SGD with the budgets stated in the main text; hardware affects wall-clock but not reported accuracies. Approximating a single eigenfunction on a fine mesh can take on the order of two hours on commodity CPUs, whereas the sequential MNIST construction at $d{=}784$ is a few minutes per coordinate on a modern accelerator when using the same iteration budget. CPU cost remains a practical bottleneck; better initialization is left to future work.

\FloatBarrier

\new{%
\section{Graph-Laplacian Pretraining Ablation}
\label{app:gl_ablation}
The optional Graph-Laplacian (GL) pretraining stage of Sec.~\ref{subsec:pretraining} fits eigenmaps on a $1000$-point subsample, distils the basis network against them for $2000$ steps, and may calibrate $T_{\mathrm{class}}$ from a held-out logistic-regression cross-entropy~\cite{4476091}. Table~\ref{tab:gl_ablation} reports a five-seed HTRU2 ablation with PIEFS-off under three settings: GL off; GL on with $T_{\mathrm{class}}$ updated from the GL probe; and GL on with $T_{\mathrm{class}}$ frozen at $0.5$. The three settings land within one standard deviation of one another on this low-dimensional benchmark.

\begin{table}[H]
  \centering
  \footnotesize
  \caption{HTRU2 test accuracy (\%, mean$\pm$std, 5 seeds). PIEFS-off with three GL settings.}
  \label{tab:gl_ablation}
  \begin{tabular}{lc}
    \toprule
    \textbf{Setting} & \textbf{Test accuracy} \\
    \midrule
    GL off (headline) & $97.90{\pm}0.15$ \\
    GL on, $T_{\mathrm{class}}$ updated & $97.94{\pm}0.11$ \\
    GL on, $T_{\mathrm{class}}$ frozen & $97.97{\pm}0.10$ \\
    \bottomrule
  \end{tabular}
\end{table}
}

\new{%
\section{Unsupervised \textsc{piefs} Ablation}
\label{app:unsup_ablation}
We run an unsupervised variant of \textsc{piefs} with $w_{\mathrm{class}}{=}0$ in~\eqref{eq:loss_function}. For this ablation we disable dynamic weighting entirely and use static weights $w_{\mathrm{orth}}{=}w_{\mathrm{mde}}{=}1$: the dynamic schedule in~\eqref{eq:loss_function_weight} gates $w_{\mathrm{mde}}$ on the class loss, which is no longer optimised when $w_{\mathrm{class}}{=}0$, so we hold the weights fixed instead. The Gram orthogonality and MDE loss \emph{terms} are otherwise unchanged. The basis network and metric heads keep the same capacity as the supervised runs (three hidden layers of width $64$). A linear probe (logistic regression, default scikit-learn settings) is fit on the train-split coordinates and evaluated on the test split. Table~\ref{tab:unsup_ablation} reports mean$\pm$std over five seeds; the NeuralEF column reuses the $K{=}16$ linear-probe rerun from Table~\ref{tab:main_results}.

\begin{table}[H]
  \centering
  \footnotesize
  \setlength{\tabcolsep}{4pt}
  \caption{Unsupervised \textsc{piefs} test accuracy (\%, mean$\pm$std, 5 seeds) with $w_{\mathrm{class}}{=}0$. NeuralEF column reports the $K{=}16$ linear-probe rerun (same as Table~\ref{tab:main_results}).}
  \label{tab:unsup_ablation}
  \resizebox{\columnwidth}{!}{%
  \begin{tabular}{l@{\hskip 6pt}cccc}
    \toprule
    \textbf{Dataset} & \textbf{off} & \textbf{diag} & \textbf{trotter} & \textbf{NeuralEF} \\
    \midrule
    Two Moons           & \new{$\mathbf{99.95{\pm}0.09}$} & \new{$80.77{\pm}2.34$} & \new{$93.07{\pm}5.28$} & \new{$88.23{\pm}0.31$} \\
    Circles             & \new{$\mathbf{100.00{\pm}0.00}$} & \new{$78.35{\pm}1.51$} & \new{$97.79{\pm}1.28$} & \new{$100.00{\pm}0.00$} \\
    HTRU2               & \new{$97.42{\pm}0.25$} & \new{$97.18{\pm}0.39$} & \new{$\mathbf{97.53{\pm}0.54}$} & \new{$97.47{\pm}0.34$} \\
    MNIST               & \new{$\mathbf{89.85{\pm}0.56}$} & \new{$78.59{\pm}1.25$} & \new{$75.39{\pm}1.01$} & \new{$82.52{\pm}0.29$} \\
    CIFAR-10 (R18 emb.) & \new{$\mathbf{79.80{\pm}0.26}$} & \new{$70.87{\pm}0.42$} & \new{$70.33{\pm}1.32$} & \new{$76.05{\pm}0.55$} \\
    \bottomrule
  \end{tabular}%
  }
\end{table}

\new{Unsupervised PIEFS-off matches or exceeds NeuralEF on four of five datasets (e.g.\ MNIST $89.9$ vs.\ $82.5$, CIFAR-10 embeddings $79.8$ vs.\ $76.1$). The learnable-metric variants, however, consistently \emph{underperform} \textsc{off} without labels: they reach lower Gram error yet lower accuracy, because with $w_{\mathrm{class}}{=}0$ the metric can trivially shrink the Dirichlet penalty by driving the volume-preserving scalings toward extreme anisotropy (we observe per-coordinate $\log\lambda$ ranges up to ${\sim}4$), so optimization capacity shifts to orthogonality at the expense of discriminative content; a per-coordinate anisotropy penalty $\mathrm{mean}\,(\log\lambda_i)^2$ partially restores them toward \textsc{off}. These observations indicate that the learnable metric requires a supervisory or geometric signal to be useful, which we leave to future work.}

\new{%
\begin{table}[H]
  \centering
  \footnotesize
  \setlength{\tabcolsep}{4pt}
  \caption{Final batch Gram error $\|C-I\|_F^2$ (lower $=$ more orthonormal; mean over seeds) for supervised and unsupervised PIEFS. Without labels the metric variants reach \emph{lower} Gram error than \textsc{off} (yet lower accuracy, Table~\ref{tab:unsup_ablation}): the volume-preserving metric trivially minimises the Dirichlet term, diverting capacity to orthogonality.}
  \label{tab:gram_errors}
  \resizebox{\columnwidth}{!}{%
  \begin{tabular}{lcccccc}
    \toprule
    & \multicolumn{3}{c}{\textbf{Supervised}} & \multicolumn{3}{c}{\textbf{Unsupervised}} \\
    \cmidrule(lr){2-4}\cmidrule(lr){5-7}
    \textbf{Dataset} & off & diag & trotter & off & diag & trotter \\
    \midrule
    Two Moons           & $0.16$ & $0.05$ & $0.05$ & $1.84$ & $0.04$ & $0.05$ \\
    Circles             & $0.12$ & $0.04$ & $0.05$ & $0.94$ & $0.25$ & $0.05$ \\
    HTRU2               & $0.09$ & $0.02$ & $0.03$ & $0.31$ & $0.02$ & $0.02$ \\
    MNIST               & $0.12$ & $0.13$ & $0.10$ & $0.22$ & $0.12$ & $0.11$ \\
    CIFAR-10 (R18 emb.) & $0.14$ & $0.12$ & $0.11$ & $0.31$ & $0.10$ & $0.11$ \\
    \bottomrule
  \end{tabular}}
\end{table}
}
}

\new{%
\section{Gram Orthogonality vs.\ Accuracy}
\label{app:gram_scatter}
Figures~\ref{fig:gram_scatter_sup}--\ref{fig:gram_scatter_unsup} plot the final Gram error $\|C-I\|_F^2$ against test accuracy for all dataset--metric combinations.
In the supervised setting (Fig.~\ref{fig:gram_scatter_sup}) all variants achieve high accuracy regardless of their Gram error, confirming that the method does not require near-perfect orthogonality to generalise well.
In the unsupervised setting (Fig.~\ref{fig:gram_scatter_unsup}) the learnable-metric variants (\textsc{diag}, \textsc{trotter}) reach \emph{lower} Gram error yet \emph{lower} accuracy than \textsc{off}: without label supervision the metric shrinks the Dirichlet penalty by driving anisotropic scalings, diverting optimisation capacity from discriminative content to orthogonality.

\begin{figure}[H]
  \centering
  \includegraphics[width=\columnwidth]{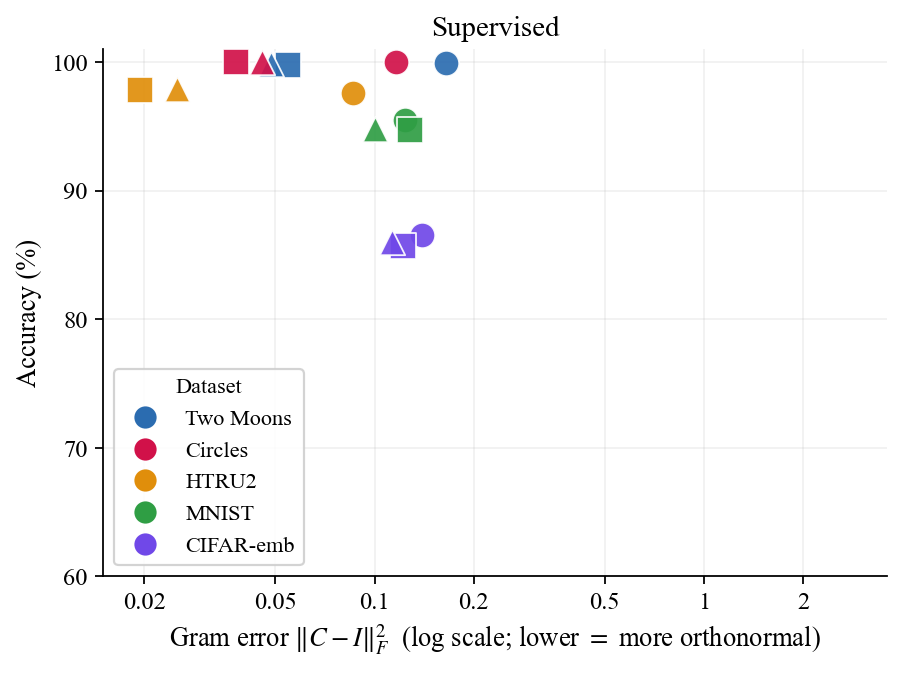}
  \caption{Supervised PIEFS: Gram error $\|C-I\|_F^2$ vs.\ test accuracy. All dataset--metric variants (circles~=~\textsc{off}, squares~=~\textsc{diag}, triangles~=~\textsc{trotter}) achieve high accuracy; orthogonality quality does not limit performance under supervision.}
  \label{fig:gram_scatter_sup}
\end{figure}

\begin{figure}[H]
  \centering
  \includegraphics[width=\columnwidth]{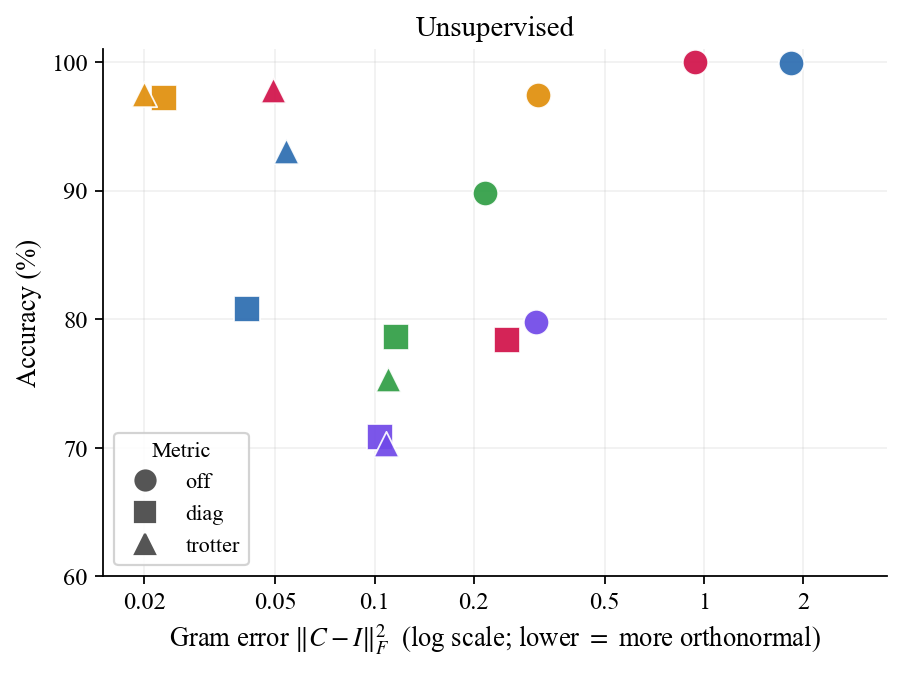}
  \caption{Unsupervised PIEFS: Gram error $\|C-I\|_F^2$ vs.\ test accuracy. Learnable-metric variants (squares, triangles) attain lower Gram error yet lower accuracy than \textsc{off} (circles), revealing the metric-degeneracy effect described in App.~\ref{app:unsup_ablation}.}
  \label{fig:gram_scatter_unsup}
\end{figure}
}

\end{document}